\documentclass[11pt,a4paper]{article}
\usepackage[hyperref]{template/emnlp-ijcnlp-2019}
\usepackage{times}
\usepackage{latexsym}
\usepackage{graphicx}
\usepackage{subcaption}
\usepackage{multirow}
\usepackage{soul}

\usepackage{url}

\aclfinalcopy 

\title{Writing habits and telltale neighbors: analyzing clinical concept\\usage patterns with sublanguage embeddings}

\author{
    Denis Newman-Griffis$^{a,b}$ \and Eric Fosler-Lussier $^a$\\
    $^a$Dept of Computer Science and Engineering, The Ohio State University, Columbus, OH\\
    $^b$Rehabilitation Medicine Dept, Clinical Center, National Institutes of Health, Bethesda, MD\\
    {\tt \{newman-griffis.1, fosler-lussier.1\} @ osu.edu}
}

\date{}

\begin{document}
\maketitle
\begin{abstract}
  Natural language processing techniques are being applied to increasingly
diverse types of electronic health records, and can benefit from
in-depth understanding of the distinguishing characteristics of medical
document types. We present a method for characterizing the usage patterns
of clinical concepts among different document types, in order to capture
semantic differences beyond the lexical level. By training concept embeddings
on clinical documents of different types and measuring the
differences in their nearest neighborhood structures, we are able to
measure divergences in concept usage while correcting for noise in embedding
learning. Experiments on the MIMIC-III corpus demonstrate that our approach
captures clinically-relevant differences in concept usage and provides an
intuitive way to explore semantic characteristics of clinical document
collections.

\end{abstract}

\section{Introduction}

Sublanguage analysis has played a pivotal role in natural language processing
of health data, from highlighting the clear linguistic differences between
biomedical literature and clinical text \cite{Friedman2002} to supporting
adaptation to multiple languages \cite{Laippala2009}. Recent studies of
clinical sublanguage have extended sublanguage study to the document type
level, in order to improve our understanding of the syntactic and lexical
differences between highly distinct document types used in modern EHR systems
\cite{Feldman2016,Gron2019}.

However, one key axis of sublanguage
characterization that has not yet been explored is how domain-specific clinical
\textit{concepts} differ in their usage patterns among different document types.
Established biomedical concepts may have multiple, often non-compositional
surface forms (e.g., ``ALS'' and ``Lou Gehrig's disease''), making them
difficult to analyze using lexical occurrence alone. Understanding how these
concepts differ between document types can not only augment recent methods for
sublanguage-based text categorization \cite{Feldman2016}, but also inform the
perennial challenge of medical concept normalization \cite{Luo2019}:
``depression'' is much easier to disambiguate if its occurrence is known to be
in a social work note or an abdominal exam.

Inspired by recent technological advances in modeling diachronic
language change \cite{Hamilton2016,Vashisth2019}, we characterize concept usage
differences within clinical sublanguages using nearest neighborhood structures
of clinical concept embeddings. We show that overlap in nearest neighborhoods
can reliably distinguish between document types while controlling for noise in
the embedding process. Qualitative analysis of these nearest neighborhoods
demonstrates that these distinctions are semantically relevant, highlighting
sublanguage-sensitive relationships between specific concepts and between
concepts and related surface forms. Our findings suggest that the structure of
concept embedding spaces not only captures domain-specific semantic
relationships, but can also identify a ``fingerprint'' of concept usage
patterns within a clinical document type to inform language understanding.

\section{Related Work}

\begin{table*}[t]
    \centering
    {\small
    \begin{tabular}{r|ccccccc}
        \multirow{2}{*}{Type}&\multirow{2}{*}{Docs}&\multirow{2}{*}{Lines}&\multirow{2}{*}{Tokens}&\multirow{2}{*}{Matches}&\multirow{2}{*}{Concepts}&\multicolumn{2}{c}{High Confidence}\\
        &&&&&&Concepts&Consistency (\%)\\
        \hline
        Case Management  &    967&    20,106&    165,608&    45,306&     557&  111& 75\\
        Consult          &     98&    15,514&     96,515&    26,109&     812&    0& --\\
        Discharge Summary& 59,652&14,480,154&104,027,364&30,840,589&   6,381&1,599& 67\\
        ECG              &209,051& 1,022,023&  7,307,381& 2,163,682&     540&   14& 56\\
        Echo             & 45,794& 2,892,069& 19,752,879& 6,070,772&   1,233&  157& 65\\
        General          &  8,301&   307,330&  2,191,618&   552,789&   2,559&    0& --\\
        Nursing          &223,586& 9,839,274& 73,426,426&18,903,892&   4,912&    2& 58\\
        Nursing/Other    &822,497&10,839,123&140,164,545&31,135,584&   5,049&   83& 60\\
        Nutrition        &  9,418&   868,102&  3,843,963& 1,147,918&   1,911&  198& 73\\
        Pharmacy         &    103&     4,887&     39,163&     8,935&     376&    0& --\\
        Physician        &141,624&26,659,749&148,306,543&39,239,425&   5,538&  122& 57\\
        Radiology        &522,279&17,811,429&211,901,548&34,433,338&   4,126&  599& 63\\
        Rehab Services   &  5,431&   585,779&  2,936,022&   869,485&   2,239&    9& 62\\
        Respiratory      & 31,739& 1,323,495&  6,358,924& 2,255,725&   1,039&    5& 63\\
        Social Work      &  2,670&   100,124&    930,674&   195,417&   1,282&    0& --\\
    \end{tabular}
    }
    \caption{Document type subcorpora in MIMIC-III. Tokenization was performed with SpaCy; Matches
             and Concepts refer to number of terminology string match instances
             and number of unique concepts embedded, respectively, using SNOMED-CT
             and LOINC vocabularies from UMLS 2017AB release. The number of high-confidence
             concepts identified for each document type is given with their mean consistency.}
    \label{tbl:doctypes}
\end{table*}

Sublanguage analysis historically focused on describing the characteristic
grammatical structures of a particular domain \cite{Friedman1986,Grishman2001,
Friedman2002}.
As methods for automated analysis of large-scale data sets have improved,
more studies have investigated lexical and semantic characteristics, such as
usage patterns of different verbs and semantic categories \cite{Denecke2014},
as well as more structural information such as document section patterns
and syntactic features \cite{Zeng2011,Temnikova2014}. The use of terminologies
to assess conceptual features of a sublanguage corpus was proposed by
\citet{Walker1986}, and \citet{Drouin2004,Gron2019} used sublanguage features
to expand existing terminologies, but large-scale characterization of concept usage
in sublanguage has remained a challenging question.

Word embedding techniques have been utilized to describe diachronic language
change in a number of recent studies, from evaluating broad changes over decades
\cite{Hamilton2016,Vashisth2019} to detecting fine-grained shifts in
conceptualizations of psychological concepts \cite{Vylomova2019}.
Embedding techniques have also been used as a mirror to analyze social biases
in language data \cite{Garg2018}. Similar to our work, \citet{Ye2018}
investigate document type-specific embeddings from clinical data as a tool for
medical language analysis. However, our approach has two significant differences:
\citet{Ye2018} used word embeddings only, while we utilize concept embeddings to
capture concepts across multiple surface forms; more importantly, their work
investigated multiple document types as a way to \textit{control} for specific usage
patterns within sublanguages in order to capture more general term similarity
patterns, while our study aims to \textit{capture} these sublanguage-specific usage
patterns in order to analyze the representative differences in language use between
different expert communities.

\section{Data and preprocessing}

We use free text notes from the MIMIC-III critical care database
\cite{Johnson2016} for our analysis. This includes approximately 2 million
text records from hospital admissions of almost 50 thousand patients to
the critical care units of Beth Israel Deaconess Medical Center over a
12-year period. Each document belongs to one of 15 document types, listed
in Table~\ref{tbl:doctypes}.

As sentence segmentation of clinical text is often optimized for specific
document types \cite{Griffis2016}, we segmented our documents at linebreaks
and tokenized using SpaCy (version 2.1.6; \citeauthor{spaCy} \citeyear{spaCy}). All tokens were lowercased, but
punctuation and deidentifier strings were retained, and no stopwords were
removed.

\section{Experiments}

\begin{figure*}[t]
    \centering
    \begin{subfigure}[b]{0.55\textwidth}
        \centering
        \includegraphics[width=\textwidth]{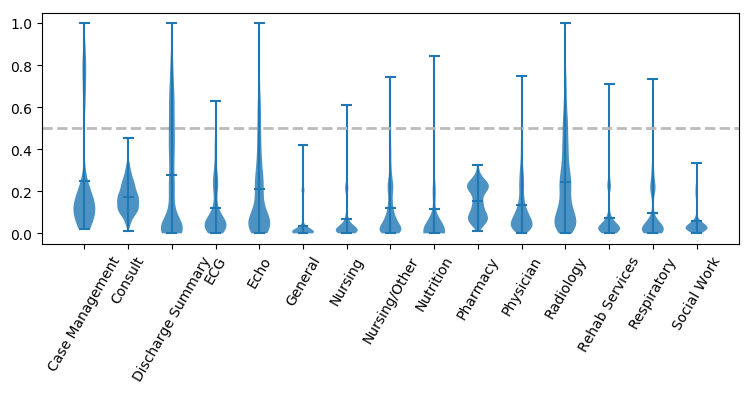}
        \caption{Self consistency by document type; line at 50\% threshold}
        \label{fig:self-consistency-all}
    \end{subfigure}
    \begin{subfigure}[b]{0.35\textwidth}
        \centering
        \includegraphics[width=\textwidth]{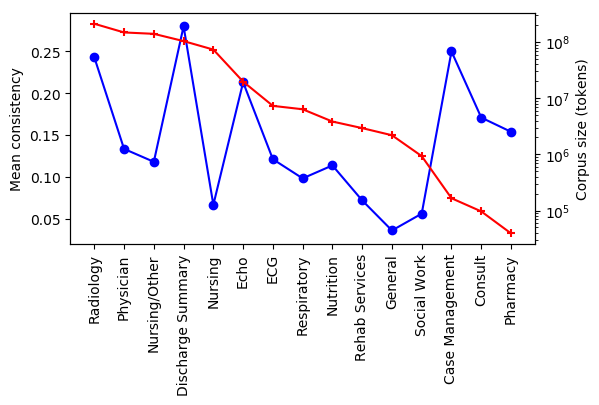}
        \caption{Self consistency compared to corpus size (log scale), with document
                 types sorted by decreasing corpus size.}
        \label{fig:self-consistency-vs-size}
    \end{subfigure}

    \caption{Distribution of self-consistency rates (i.e., overlap in nearest
             neighbors between replicate embeddings of the same concept) among
             MIMIC document types.}
    \label{fig:self-consistency}
\end{figure*}

Methods for learning clinical concept representations have proliferated in
recent years \cite{Choi2016KDD,Mencia2016,Phan2019}, but often require
annotations in forms such as billing codes or disambiguated concept mentions.
These annotations may be supplied by human experts such as medical coders,
or by adapting medical NLP tools such as MetaMap \cite{Aronson2010} or
cTAKES \cite{Savova2010} to perform concept recognition \cite{DeVine2014}.

For investigating potentially divergent usage patterns of
clinical concepts, these strategies face serious limitations: the full
diversity of MIMIC data has not been annotated for concept identifiers,
and the statistical biases of trained NLP tools may suppress underlying
differences in automatically-recognized concepts. We therefore take a
distant supervision approach, using JET
\cite{Newman-Griffis2018b}.  JET uses a sliding context window to jointly
train embedding models for
words, surface forms, and concepts, using a log-bilinear objective with
negative sampling and shared embeddings for context words. It leverages
known surface forms from a terminology as a source of distant
supervision: each occurrence of any string in the terminology is treated
as a weighted training instance for each of the concepts that string can
represent. As terminologies are typically many-to-many maps between 
surface forms and concepts, this generally leads to a unique set of contexts
being used to train the embedding of each concept, though any individual
context window may be used as a sample for training multiple concepts.
We constrain the scope of our analysis to only
concepts and strings from SNOMED-CT and LOINC,\footnote{
    We used the versions distributed in the 2017AB release of the UMLS \cite{Bodenreider2004}.
} two popular high-coverage clinical vocabularies.

\begin{figure*}[t]
    \centering
    \begin{subfigure}[b]{0.32\textwidth}
        \centering
        \includegraphics[width=\textwidth]{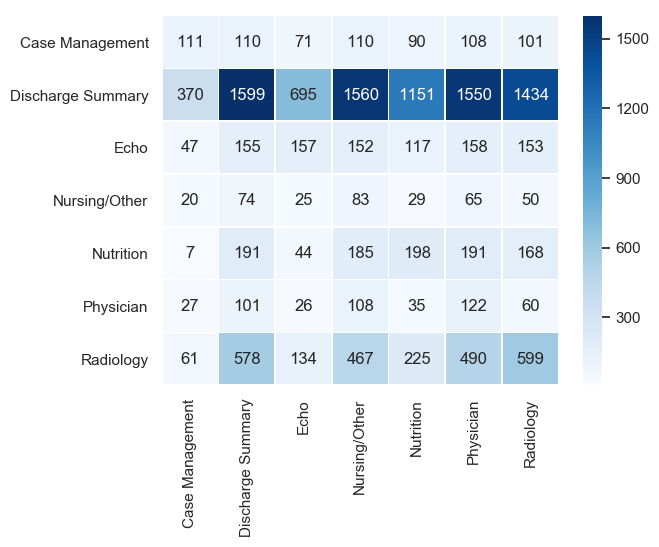}
        \caption{Number of concepts analyzed}
        \label{fig:nbr-analysis-counts}
    \end{subfigure}
    \begin{subfigure}[b]{0.32\textwidth}
        \centering
        \includegraphics[width=\textwidth]{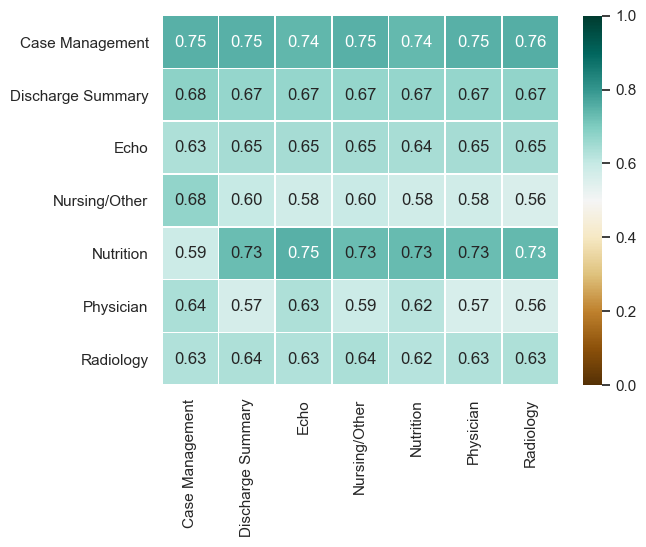}
        \caption{Reference set self-consistency}
        \label{fig:nbr-analysis-src}
    \end{subfigure}
    \begin{subfigure}[b]{0.32\textwidth}
        \centering
        \includegraphics[width=\textwidth]{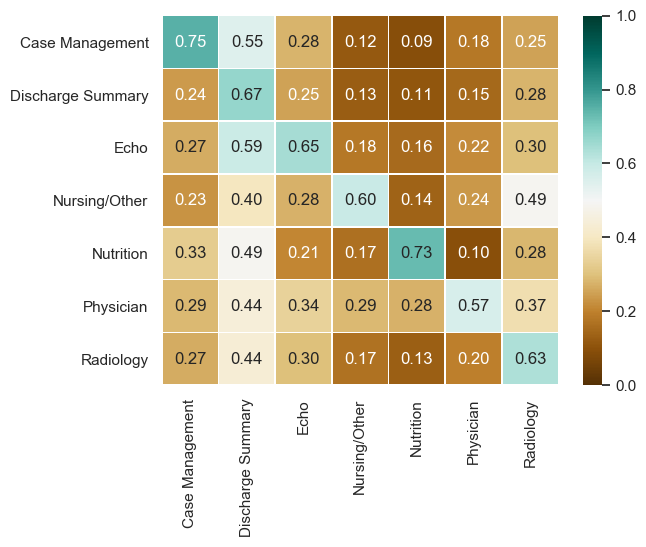}
        \caption{Comparison set self-consistency}
        \label{fig:nbr-analysis-trg}
    \end{subfigure}
    \begin{subfigure}[b]{0.35\textwidth}
        \centering
        \includegraphics[width=\textwidth]{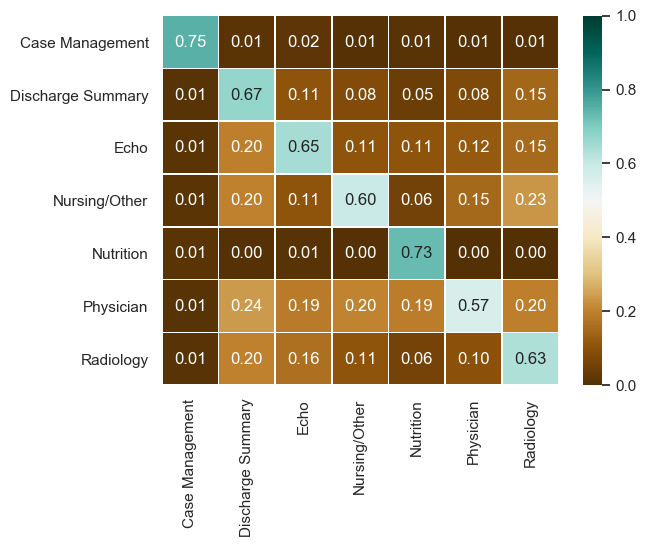}
        \caption{Cross-type consistency}
        \label{fig:nbr-analysis-cross}
    \end{subfigure}
    \begin{subfigure}[b]{0.35\textwidth}
        \centering
        \includegraphics[width=\textwidth]{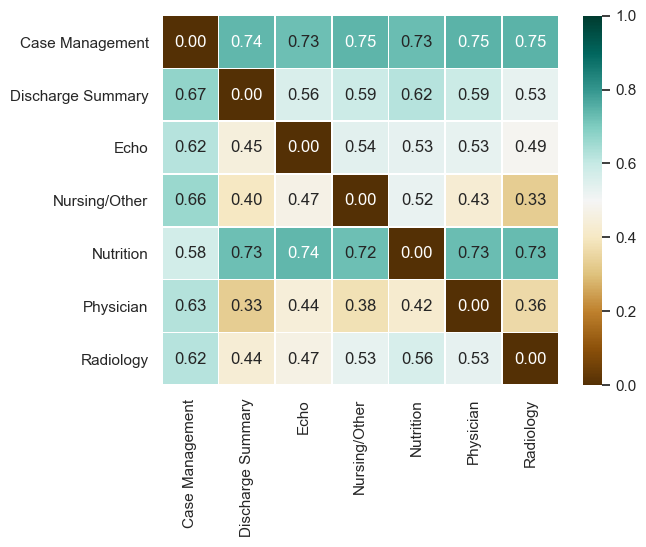}
        \caption{Consistency deltas}
        \label{fig:nbr-analysis-deltas}
    \end{subfigure}

    \caption{Comparison of concept neighborhood consistency statistics across
             document types, using high-confidence concepts from the reference
             type. Figure~\ref{fig:nbr-analysis-counts} provides the number of
             concepts shared between the high-confidence reference set and the
             comparison set. All values are the mean of the consistency
             distribution calculated over all concepts analyzed for the
             document type pair.}
    \label{fig:nbr-analysis}
\end{figure*}

\begin{figure*}[t]
    \centering
    \begin{subfigure}[b]{0.45\textwidth}
        \centering
        \includegraphics[width=\textwidth]{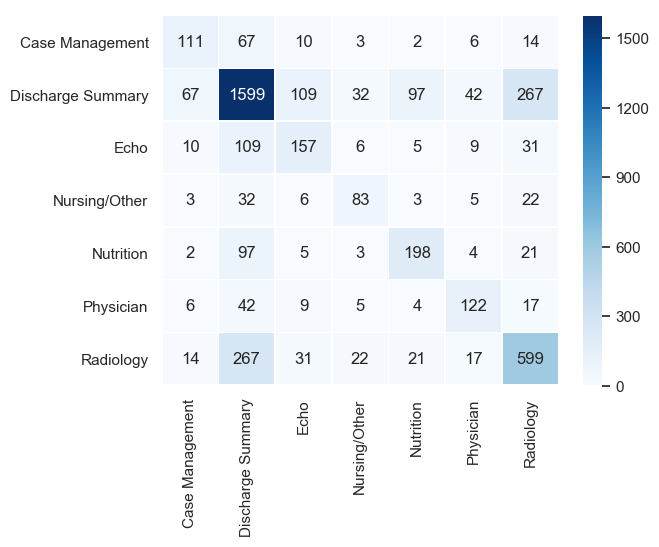}
        \caption{Number of concepts analyzed}
        \label{fig:nbr-analysis-counts-shc}
    \end{subfigure}
    \begin{subfigure}[b]{0.45\textwidth}
        \centering
        \includegraphics[width=\textwidth]{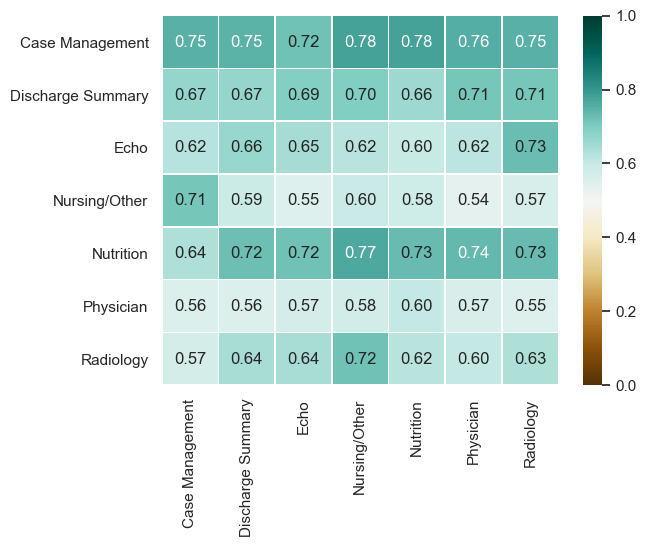}
        \caption{Self-consistency}
        \label{fig:nbr-analysis-src-shc}
    \end{subfigure}
    \begin{subfigure}[b]{0.45\textwidth}
        \centering
        \includegraphics[width=\textwidth]{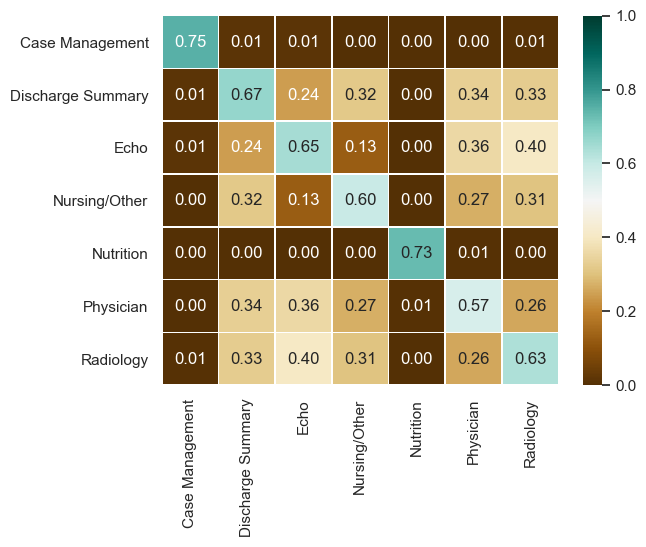}
        \caption{Cross-type consistency}
        \label{fig:nbr-analysis-cross-shc}
    \end{subfigure}
    \begin{subfigure}[b]{0.45\textwidth}
        \centering
        \includegraphics[width=\textwidth]{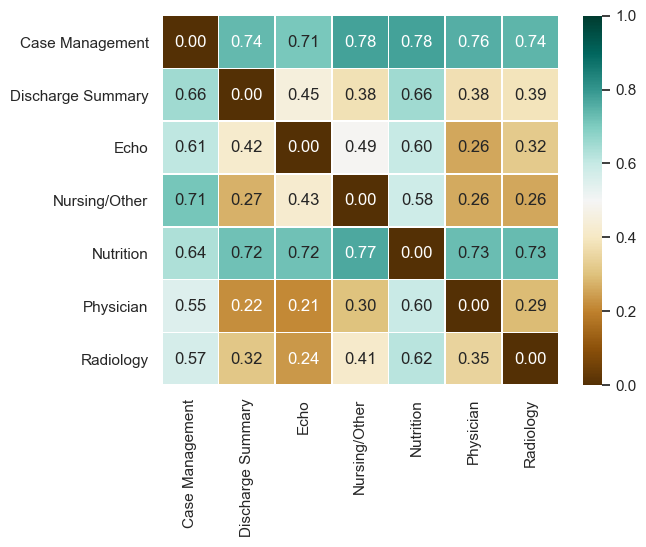}
        \caption{Consistency deltas}
        \label{fig:nbr-analysis-deltas-shc}
    \end{subfigure}

    \caption{Concept neighborhood consistency statistics, restricted to concepts
             that are high-confidence in both reference and comparison sets. In
             this case, reference self-consistency and target self-consistency
             are symmetric, so only reference self-consistency is presented in
             Figure~\ref{fig:nbr-analysis-src-shc}.}
    \label{fig:nbr-analysis-shc}
\end{figure*}

\subsection{Identifying concepts for comparison}

For each document type, we concatenate all of its documents (maintaining
linebreaks), identify all occurrences of SNOMED-CT and LOINC strings in each
line, and use these occurrences to train word, term, and concept embeddings
with JET. Due to the size of our subcorpora, we used a window size of 5,
minimum frequency of 5, embedding dimensionality of 100, initial learning rate
of 0.05, and 10 iterations over each corpus.

Prior research has noted instability of nearest neighborhoods in multiple embedding
methods \cite{Wendlandt2018}.  We therefore train 10 sets of embeddings from each
of our subcorpora, each using the same hyperparameter settings but a different
random seed.  We then use all 10 replicates from each subcorpus in our analyses,
in order to control for variation in nearest neighborhoods introduced by random
initialization and negative sampling.  To evaluate
the baseline reliability of concept embedding neighborhoods from each subcorpus, we
calculated per-concept consistency by measuring, over all pairs of embedding sets
within the 10 replicates, the average set membership overlap
between the top 5 nearest neighbors by cosine similarity for each concept
embedding.\footnote{
    We chose five nearest neighbors for our analyses based on qualitative review
    of neighborhoods for concepts within different document types.  We found
    nearest neighborhoods for concept embeddings to vary more than for word
    embeddings, often introducing noise beyond the top five nearest neighbors;
    we therefore set a conservative baseline for reliability by focusing on the
    closest and most stable neighbors.  However, using 10 neighbors,
    as \citet{Wendlandt2018} did, or more could yield different qualitative
    patterns in document type comparisons and bears exploration.
} As shown in Figure~\ref{fig:self-consistency-all}, these consistency
scores vary widely both within and between document types, with some document
types producing no concept embeddings with consistency over 40\%. Interestingly,
as illustrated in Figure~\ref{fig:self-consistency-vs-size}, there is no linear
relationship between log corpus size and mean concept consistency
($R^2\approx0.011$), suggesting that low consistency is not solely due to
limited training data.

To mitigate concerns about the reliability of embeddings for comparison,
a set of \textbf{high-confidence concepts} is identified for each document type by
retaining only those with a self-consistency of at least 50\%;
Table~\ref{tbl:doctypes} includes the number of high-confidence concepts
identified and the mean consistency among this subset.\footnote{
    We found in our analysis that most concept consistency numbers clustered
    roughly bimodally, between 0-30\% or 60-90\%; this is reflected at a coarse level
    in the overall distributions in Figure~\ref{fig:self-consistency-all}.
    Varying the threshold outside of these ranges did not have a significant
    impact on the number of concepts retained; the 50\% threshold was chosen
    for simplicity. With larger corpora, yielding higher concept coverage, a
    higher threshold could be chosen for a stricter analysis.
} These embeddings capture
reliable concept usage information for each document type, and form the basis
of our comparative analysis.

\subsection{Cross-corpus analysis}

Our key question is what concept embeddings reveal about clinical concept usage
\textit{between} document types. To maintain a sufficient sample size, we
restrict our comparison to the 7 document types with at least 50 high confidence
concepts: \textit{Case Management}, \textit{Discharge Summary}, \textit{Echo},
\textit{Nursing/Other}, \textit{Nutrition}, \textit{Physician}, and
\textit{Radiology}. \textit{Physician}, \textit{ECG}, and \textit{Nursing} were
also used by \citet{Feldman2016} for their lexicosyntactic analysis, although
they combined \textit{Nursing} documents (longer narratives) and
\textit{Nursing/Other} (which tend to be much shorter) into a single set, while
we retain the distinction. Interestingly, the fourth type they analyzed,
\textit{ECG}, produced only 14 high-confidence
concepts in our analysis, suggesting high semantic variability despite the large
number of documents.

As learned concept sets differ between document types, the first step for
comparing a document type pair is to identify the set of concepts embedded for
both. For reference type $A$ and comparison type $B$, we identify
high-confidence concepts from $A$ that are also present in $B$, and calculate
four distributions using this shared set:

\textbf{Reference consistency:} self-consistency across each of the shared
concepts, using only other shared concepts to identify nearest
neighborhoods in embeddings for the reference set.

\textbf{Comparison consistency:} self-consistency of each shared concept in
embeddings for the comparison document type, again
using only shared concepts for neighbors. As the shared
set is based on high-confidence concepts from the reference set,
this is not symmetric with reference consistency (as
the high-confidence sets may differ).

\textbf{Cross-type consistency:} average consistency for each shared concept
calculated over all pairs of replicates (i.e., comparing the nearest
neighbors of all 10 reference embedding sets to the nearest neighbors
in all 10 comparison embedding sets).

\textbf{Consistency deltas:} the difference, for each shared concept, between
its reference self-consistency and its cross-type consistency. This
provides a direct evaluation of how distinct the concept usage is
between two document types, where a high delta indicates highly
distinct usage.

\begin{table*}[t]
    \centering
    {\small
    \renewcommand\arraystretch{1.2}
    \begin{tabular}{p{0.15\linewidth}|p{0.21\linewidth}|p{0.21\linewidth}|p{0.21\linewidth}}
        \multirow{1}{*}{Query}
            &\multicolumn{1}{c}{Discharge Summary}
            &\multicolumn{1}{c}{Nursing/Other}
            &\multicolumn{1}{c}{Radiology}\\
        \hline
        \multirow{10}{\linewidth}{Diabetes Mellitus (C0011849)}
            &Diabetes (C0011847)
            &Gestational Diabetes (C0085207)
            &Poorly controlled (C3853134)
            \\

            &Type 2 (C0441730)
            &A2 immunologic symbol (C1443036)
            &Insulin (C0021641)
            \\

            &Type 1 (C0441729)
            &Diabetes Mellitus, Insulin-Dependent (C0011854)
            &Diabetes Mellitus, Insulin-Dependent (C0011854)
            \\

            &Gestational Diabetes (C0085207)
            &Factor V (C0015498)
            &Diabetes Mellitus, Non-Insulin-Dependent (C0011860)
            \\

            &Diabetes Mellitus, Insulin-Dependent (C0011854)
            &A1 immunologic symbol (C1443035)
            &Stage level 5 (C0441777)
            \\

        \hline
        \multicolumn{1}{c}{}&\multicolumn{1}{c}{Discharge Summary}&\multicolumn{1}{c}{Echo}&\multicolumn{1}{c}{Radiology}\\
        \hline
        \multirow{10}{\linewidth}{Mental state (C0278060)$\dagger$}
            &Coherent (C4068804)
            &Donor:Type:Point in time:\^{}Patient:Nominal (C3263710)
            &Mental status changes (C0856054)
            \\

            &Confusion (C0009676)
            &Donor person (C0013018)
            &Abnormal mental state (C0278061)
            \\

            &Respiratory status:-:Point in time:\^{}Patient:- (C2598168)
            &Respiratory arrest (C0162297)
            &Level of consciousness (C0234425)
            \\

            &Respiratory status (C1998827)
            &Organ donor:Type:Point in time:\^{}Donor:Nominal (C1716004)
            &Level of consciousness:Find:Pt:\^{}Patient:Ord (C4050479)
            \\

            &Abnormal mental state (C0278061)
            &Swallowing G-code (C4281783)
            &Mississippi (state) (C0026221)
            \\
    \end{tabular}
    }
    \caption{5 nearest neighbor concepts to \textit{Diabetes Mellitus} and
             \textit{Mental state} from 3 high-confidence document types,
             averaging cosine similarities across all replicate embedding sets
             within each document type. $\dagger$The two nearest neighbors to
             \textit{Mental state} for all three document types were two LOINC
             codes using the same ``mental status'' string; they are omitted
             here for brevity.}
    \label{tbl:concept-neighbors}
\end{table*}

Mean values for these distributions are provided for each pair of our 7
document types in Figure~\ref{fig:nbr-analysis}.
Comparing Figures~\ref{fig:nbr-analysis-src} and \ref{fig:nbr-analysis-trg}, it
is clear that high-confidence concepts for one document type are typically not
high-confidence for another. Most document type pairs show fairly strong
divergence, with deltas ranging from 30-60\%. \textit{Physician} notes have
comparatively high cross-set consistency of around 20\% for their high-confidence
concepts, likely reflecting the all-purpose nature of these documents, which
include patient history, medications, vitals, and detailed examination notes.
Interestingly, \textit{Case Management} and \textit{Nutrition} are starkly
divergent from other document types, with near-zero cross-set consistency and
comparatively high self-consistency of over 70\% in the compared concept sets,
despite a relatively high overlap between their high-confidence sets and concepts
learned for other document types.

In order to control for the low overlap between high-confidence sets in different
document types, we also re-ran our consistency analysis restricted to only concepts
that are high-confidence in \textit{both} the reference and comparison sets. As
shown in Figure~\ref{fig:nbr-analysis-shc}, this yields considerably smaller
concept sets for comparison, with single-digit overlap for 18/42 non-self pairings.
Cross-set consistency increases somewhat, most significantly for
pairings involving \textit{Physician} or \textit{Radiology}; however, no consistency
delta falls below 20\% for any non-self pair, indicating that concept neighborhoods
remain distinct even within high-confidence sets.

\subsection{Qualitative neighborhood analysis}
\label{sec:qualitative-nn-analysis}

Analysis of neighborhood consistency enables measuring divergence in the
contextual usage patterns of clinical concepts; however, this divergence
could be due to spurious or semantically uninformative correlations instead
of clinically-relevant distinctions in concept similarities. To confirm that
our methodology captures informative distinctions in concept usage, we
qualitatively review example neighborhoods.  To mitigate variability of
nearest neighborhoods in embedding spaces, we identify a concept's
\textit{qualitative} nearest neighbors for a given document type by calculating
its pairwise cosine distance vectors for all 10 replicates in that document
type and taking the $k$ neighbors with lowest average distance.

As with our consistency analyses, we focus on the neighborhoods of
high-confidence concepts, although we do not filter the neighborhoods
themselves. Of all high-confidence concepts identified in our embeddings, only
two were high-confidence in 5 different document types, and these were highly
generic concepts: \textit{Interventional procedure} (C0184661) and a
corresponding LOINC code (C0945766). Seven concepts were high-confidence for
4 document types; of these, two were generic procedure concepts, two were
concepts for the broad gastrointestinal category, and three were versions of
body weight. For a diversity of concepts, we therefore turned to the 75 concepts
that were high-confidence within 3 document types. We reviewed each of these
concepts, and describe our findings for three of the most broadly
clinically-relevant below.

\textbf{Diabetes Mellitus (C0011849)} \textit{Diabetes Mellitus} (search strings:
``diabetes mellitus'' and ``diabetes mellitus dm'') was
high-confidence in \textit{Discharge Summary}, \textit{Nursing/Other},
and \textit{Radiology} document types; Table~\ref{tbl:concept-neighbors} gives
the top 5 neighbors from each type. These neighbors are
semantically consistent across document types: more specific diabetes-related
concepts, related biological factors; continuing down the nearest neighbors
list yields related symptoms and comorbidities such as
\textit{Irritable Bowel Syndrome} (C0022104) and \textit{Gastroesophageal
reflux disease} (C0017168).

\begin{figure}[t]
    \centering
    \includegraphics[width=0.4\textwidth]{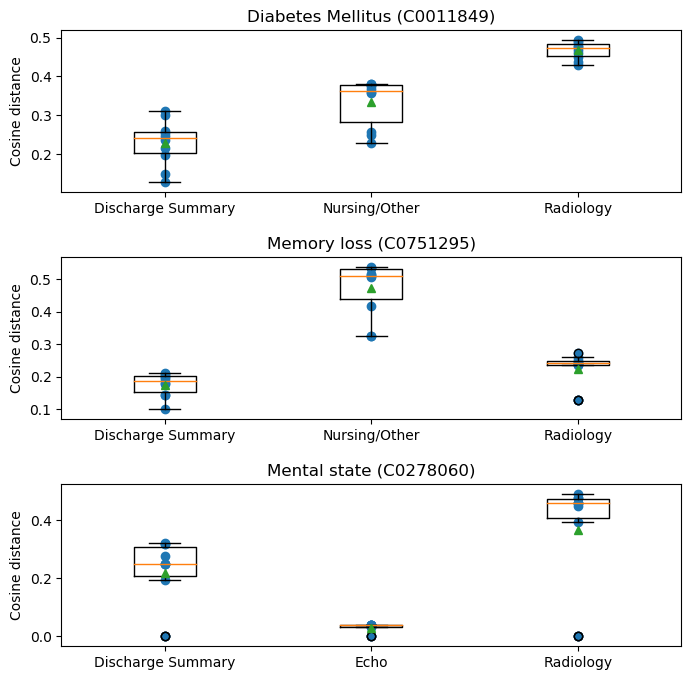}
    \caption{Cosine distance distribution of three concepts to their 10 nearest
             neighbors, averaged across document type replicate embeddings.}
    \label{fig:concept-neighbor-distances}
\end{figure}

\textbf{Memory loss (C0751295)} \textit{Memory loss} (search string: ``memory loss'')
was also high-confidence
in \textit{Discharge Summary}, \textit{Nursing/Other}, and \textit{Radiology}
documents. For brevity, its nearest neighbors are omitted from
Table~\ref{tbl:concept-neighbors}, as there is little variation
among the top 5. However, the next neighbors (at only slightly greater cosine
distance) vary considerably across document types, while remaining highly
consistent within each individual type. In
\textit{Discharge Summary}, more high-level concepts related to overall function
emerge, such as \textit{Functional status} (C0598463), \textit{Relationships}
(C0439849), and \textit{Rambling} (C4068735). \textit{Radiology} yields more
symptomatically-related neighbors: \textit{Aphagia} (C0221470) is present in
both, but \textit{Radiology} includes \textit{Disorientation} (C0233407),
\textit{Delusions} (C0011253), and \textit{Gait, Unsteady} (C0231686).
Finally, \textit{Nursing/Other} finds concepts more related to daily life, such
as \textit{Cigars} (C0678446) and \textit{Multifocals} (C3843228), though at
a greater cosine distance than the other document types
(Figure~\ref{fig:concept-neighbor-distances}).

\textbf{Mental state (C0278060)} \textit{Mental state} (search strings:
``mental status'', ``mental state'') was high-confidence in \textit{Discharge Summary},
\textit{Echo}, and \textit{Radiology}, and highlighted an unexpected
consequence of relying on the Distributional Hypothesis \cite{Harris1954}
for semantic characterization in sublanguage-specific corpora.
The top 5 nearest neighbors
(excluding two trivial LOINC codes for the same concept, also using the
``mental status'' search string) are given in Table~\ref{tbl:concept-neighbors}.
In \textit{Discharge Summary} documents, ``mental status'' is typically referred
to in detailed patient narratives, medication lists, and the like, and this
yields semantically-reasonable nearest neighbors such as \textit{Confusion}
(C0009676) and \textit{Coherent} (C4068804).

\begin{table*}[t]
    \centering
    {\small
    \begin{tabular}{l|lll}
        {\centering Query}&{\centering Discharge Summary}&{\centering Nutrition}&{\centering Case Management}\\
        \hline
        \multirow{5}{*}{Community (C0009462)}
            &Community&Dilute&Substance\\
            &Health center&Social work&Monitoring\\
            &Acquired&Surgical site&Somewhat\\
            &Residence&In situ&Hearing\\
            &Nursing facility&Nephritis&Speech\\
        \hline
        \multicolumn{1}{c}{}&\multicolumn{1}{c}{Discharge Summary}&\multicolumn{1}{c}{Echo}&\multicolumn{1}{c}{ECG}\\
        \hline
        \multirow{5}{*}{ECG (C0013798)}
            &ECG&ECG&ECG\\
            &EKG&Exercise&Physician\\
            &Sinus tachycardia&Stress&Last\\
            &Sinus bradycardia&Fair&No change\\
            &Right bundle branch block&Specific&Abnormal\\
        \hline
        \multicolumn{1}{c}{}&\multicolumn{1}{c}{Discharge Summary}&\multicolumn{1}{c}{Echo}&\multicolumn{1}{c}{Radiology}\\
        \hline
        \multirow{5}{*}{Blood pressure (C0005823)}
            &Blood pressure&Blood pressure&Blood pressure\\
            &Heart rate&Heart rate&Heart rate\\
            &Pressure&Rate&Rate\\
            &Systolic blood pressure&Exercise&Method\\
            &Rate&Stress&Exercise\\
    \end{tabular}
    }
    \caption{5 nearest neighbor surface forms to three frequent clinical
             concepts, across document types for which they are high-confidence.}
    \label{tbl:term-neighbors}
\end{table*}

In \textit{Echo} documents, however,
``mental status'' occurs most frequently within an ``Indication'' field
of the ``PATIENT/TEST INFORMATION'' section. Two common patterns emerge in
``Indication'' texts: references to altered or reduced mental status, or
patients who are vegetative and being evaluated for organ donor eligibility.
Though ``mental status'' and ``organ donor'' do not co-occur, their consistent
occurrence in the same contextual structures leads to extremely similar embeddings
(see Figure~\ref{fig:concept-neighbor-distances}). A
similar issue occurs in \textit{Radiology} notes, where the ``MEDICAL CONDITION''
section includes several instances of elderly patients presenting with either
hypothermia or altered mental status; as a result, two hypothermia concepts
(C1963170 and C0020672) are in the 10 nearest neighbors to \textit{Mental state}.

Results from \textit{Radiology} also highlight one limitation of distant supervision
for learning concept embeddings: as the word ``state'' is polysemous,
including a geopolitical entity,
geographical concepts such as \textit{Mississippi} (C0026221) end up with similar
embeddings to \textit{Mental state}. A similar issue occurs in the neighbors for
\textit{Memory loss}; due to string polysemy, the concept \textit{CIGAR string -
sequence alignment} (C4255278) ends up with a similar embedding to \textit{Cigars}
(C0678446).

\subsection{Nearest surface form embeddings}

As JET learns embeddings of concepts and their surface forms jointly in a single
vector space, we also analyzed the surface forms embeddings nearest to different
concepts. This enabled us both to evaluate the semantic congruence of surface
form and concept embeddings, and to further delve into corpus-specific contextual
patterns that emerge in the vector space. As with our concept neighborhood
analysis, for each of our 10 replicate embeddings in each document type, we
calculated the cosine distance vector from each high-confidence concept to all of
the term embeddings in the same replicate, and then averaged these distance
vectors to identify neighbors robust to embedding noise.
Table~\ref{tbl:term-neighbors} presents surface form neighbors identified for
three high-confidence clinical concepts chosen for clinical relevance and
wide usage; these concepts are discussed in the following paragraphs.

\textbf{Blood pressure (C0005823)} \textit{Blood pressure} is high-confidence in
\textit{Discharge Summary}, \textit{Echo}, and \textit{Radiology} documents. It
is a key concept that is measured frequently in various settings; intuitively, it
is a sufficiently core concept that it should exhibit little variance. Its
neighbor surface forms indeed indicate fairly consistent use across the three
document types, referencing both related measurements (``heart rate'') and related
concepts (``exercise'' and ``stress'').

\textbf{Echocardiogram (C0013798)} \textit{Echocardiogram} is high-confidence in
\textit{Discharge Summary}, \textit{Echo} (detailed summaries and interpretation
written after the ECG), and \textit{ECG} (technical notes taken during the
procedure) documents. ECGs are common, and are performed for various
purposes and discussed in varying detail. Interestingly, neighbor surface forms in
\textit{Discharge Summary} embeddings reflect specific pathologies, potentially
capturing details determined post diagnosis and treatment. In \textit{Echo}
embeddings, the neighbors are more general surface forms evaluating the findings
(``fair'') and relevant history/symptoms that led to the ECG (``exercise'',
``stress''). \textit{ECG} embeddings reflect their more technical nature,
with surface forms such as ``no change'' and ``abnormal'' yielding
high similarity.

\textbf{Community (C0009462)} \textit{Community} is a very broad concept and a
common word, and is discussed primarily in documents concerned with whole-person
health; it is high confidence in \textit{Discharge Summary}, \textit{Nutrition},
and \textit{Case Management} documents. Each of these document types reflects
different usage patterns. The nearest surface forms in \textit{Discharge Summary}
embeddings reflect a focus on living conditions, referring to ``health center'',
``residence'', and ``nursing facility''. In \textit{Nutrition} documents,
\textit{Community} is discussed primarily in terms of ``community-acquired
pneumonia'', likely leading to more treatment-oriented neighbor surface forms.
Finally, in \textit{Case Management} embeddings, nearby surface forms reflect
discussion of specific risk factors or resources (``substance'', ``monitoring'')
to consider in maintaining the
patient's health and responding to their specific needs (e.g., ``hearing'', ``speech'').

\section{Discussion}

We have shown that concept embeddings learned from different clinical document
type corpora reveal characteristics of how clinical concepts are used in
different settings. This suggests that sublanguage-specific embeddings can help
profile distinctive usage patterns for text categorization, offering greater
specificity than latent topic distributions while not relying on potentially brittle
lexical features. In addition, such profiles could also assist with
concept normalization by providing more-informed prior probability
distributions for medical vocabulary senses that are conditioned on the document
or section type that they occur in.

A few limitations of our study are important to note. The embedding method we
chose offers flexibility to work with arbitrary corpora and vocabularies, but
its use of distant supervision introduces some undesirable noise. The example
given in Section~\ref{sec:qualitative-nn-analysis} of the similar embeddings
learned for the concept \textit{cigars} and the concept of the CIGAR string
in genomic sequence editing illustrates the downside of not leveraging
disambiguation techniques to filter out noisy matches. On the other hand, our
restriction to strings from SNOMED-CT and LOINC provided a high-quality set of
strings intended for clinical use, but also removed many potentially
helpful strings from consideration. For example, the UMLS also includes the
non-SNOMED/LOINC strings ``diabetes'' and ``diabete mellitus'' [\textit{sic}] for
\textit{Diabetes Mellitus} (C0011849), both of which occur frequently in MIMIC
data. Misspellings are also common in clinical data; leveraging well-developed
technologies for clinical spelling correction would likely increase
the coverage and confidence of sublanguage concept embeddings.

At the same time, the low volume of data analyzed in many document types
introduces its own challenges for the learning process. First, though JET can
in principle learn embeddings for every concept in a given terminology, this is
predicated on the relevant surface forms appearing with sufficient frequency.
For a small document sample, many such surface forms that would otherwise be
present in a larger sample will either be missing entirely or insufficiently
frequent, leading to effectively ``missed'' concepts.  While we are not aware
of another concept embedding method compatible with arbitrary unannotated
corpora that could help avoid these issues, some strategies could be used to
reduce the potential impact of both training noise and low sample sizes.
One approach, which might also help improve concept consistency in the document
types that yielded few or no high-confidence concepts, would be pretraining
a shared base embedding on a large corpus such as PubMed abstracts, which
could then be tuned on each document type-specific subcorpus.  While this could
introduce its own noise in terms of the differences between biomedical
literature language and clinical language \cite{Friedman2002}, it could help
control for some degree of sampling error and provide a linguistically-motivated
initialization for the concept embedding models.

Finally, as we observed with \textit{Mental state} (C0278060), relying on similarity
in contextual patterns can lead to capturing more \ul{corpus}-specific features
with embeddigns, as opposed to \ul{(sub)language}-specific features, as target
corpora become smaller and more
homogeneous. If a particular concept or set of concepts are always used within
the same section of a document, or in the same set phrasing, the ``similarity''
captured by organization of an embedding space will be more informed by this
writing habit endemic to the specific corpus than by clinically-informed semantic
patterns that can generalize to other corpora.

\section{Conclusion}

Analyzing nearest neighborhoods in embedding spaces has become a powerful tool
in studying diachronic language change.  We have described how the same
principles can be applied to sublanguage analysis, and
demonstrated that the structure of concept embedding spaces captures
distinctive and relevant semantic
characteristics of different clinical document types.  This offers a valuable
tool for sublanguage characterization, and a promising avenue for developing
document type ``fingerprints'' for text categorization and knowledge-based
concept normalization.

\section*{Acknowledgments}

The authors gratefully thank Guy Divita for helpful feedback
on early versions of the manuscript.
This research was supported by the Intramural Research Program of the
National Institutes of Health and the US Social Security Administration.

\bibliography{references}
\bibliographystyle{template/acl_natbib}

\end{document}